\documentclass[letterpaper]{article} % DO NOT CHANGE THIS
\usepackage{aaai24}  % DO NOT CHANGE THIS
\usepackage{times}  % DO NOT CHANGE THIS
\usepackage{helvet}  % DO NOT CHANGE THIS
\usepackage{courier}  % DO NOT CHANGE THIS
\usepackage[hyphens]{url}  % DO NOT CHANGE THIS
\usepackage{graphicx} % DO NOT CHANGE THIS
\usepackage{natbib}  % DO NOT CHANGE THIS AND DO NOT ADD ANY OPTIONS TO IT
\usepackage{caption} % DO NOT CHANGE THIS AND DO NOT ADD ANY OPTIONS TO IT
\usepackage{algorithm}
\usepackage{algorithmic}
\usepackage{multirow}
\usepackage{amsmath}

\usepackage{amsmath}
\usepackage{amsfonts}
\usepackage{booktabs}
\usepackage{multirow}
\usepackage{multicol}                 % 合并多列
\usepackage{multirow}                % 合并多行
\usepackage{float}                    % 图片浮动
\usepackage{makecell}                 % 三线表-竖线
\usepackage{booktabs}  
\usepackage{bbding}
\usepackage{newfloat}
\usepackage{listings}
\DeclareCaptionStyle{ruled}{labelfont=normalfont,labelsep=colon,strut=off} % DO NOT CHANGE THIS
\floatstyle{ruled}
\newfloat{listing}{tb}{lst}{}
\floatname{listing}{Listing}
\title{MapExpert: Online HD Map Construction with Simple and Efficient\\Sparse Map Element Expert}

\author{
    Dapeng Zhang\textsuperscript{\rm 1}, Dayu Chen\textsuperscript{\rm 2}, Peng Zhi\textsuperscript{\rm 1}\footnotemark[1], Yinda Chen\textsuperscript{\rm 3}, Zhenlong Yuan\textsuperscript{\rm 4},\\ Chenyang Li\textsuperscript{\rm 1}, Sunjing\textsuperscript{\rm 1}, Rui Zhou\textsuperscript{\rm 1}, Qingguo Zhou\textsuperscript{\rm 1}\thanks{Corresponding author.}  
}
\affiliations{
    \textsuperscript{\rm 1}School of Information Science \& Engineering, Lanzhou University, China\\
    \textsuperscript{\rm 2}Smart (Shanghai) Robotics Technology Co., Ltd., China\\
    \textsuperscript{\rm 3}School of Information Science and Technology, University of Science and Technology of China, China\\
    \textsuperscript{\rm 4}Institute of Computing Technology, Chinese Academy of Sciences, China\\
    
    \{zhangdp22, zhip13, lchengyang2024, sjing2023, zr, zhouqg\}@lzu.edu.cn, 
    cassidy.chen@smart.com, 
    cyd0806@mail.ustc.edu.cn,
    yuanzhenlong21b@ict.ac.cn
}

\usepackage{bibentry}
\begin{document}

\maketitle

\begin{abstract}
Constructing online High-Definition (HD) maps is crucial for the static environment perception of autonomous driving systems (ADS). Existing solutions typically attempt to detect vectorized HD map elements with unified models; however, these methods often overlook the distinct characteristics of different non-cubic map elements, making accurate distinction challenging. To address these issues, we introduce an expert-based online HD map method, termed MapExpert. MapExpert utilizes sparse experts, distributed by our routers, to describe various non-cubic map elements accurately. Additionally, we propose an auxiliary balance loss function to distribute the load evenly across experts. Furthermore, we theoretically analyze the limitations of prevalent bird's-eye view (BEV) feature temporal fusion methods and introduce an efficient temporal fusion module called Learnable Weighted Moving Descent. This module effectively integrates relevant historical information into the final BEV features. Combined with an enhanced slice head branch, the proposed MapExpert achieves state-of-the-art performance and maintains good efficiency on both nuScenes and Argoverse2 datasets.
\end{abstract}

% Uncomment the following to link to your code, datasets, an extended version or similar.
%
% \begin{links}
%     \link{Code}{https://aaai.org/example/code}
%     \link{Datasets}{https://aaai.org/example/datasets}
%     \link{Extended version}{https://aaai.org/example/extended-version}
% \end{links}

\begin{figure*}[t]
\centering
\includegraphics[width=1.00\textwidth]{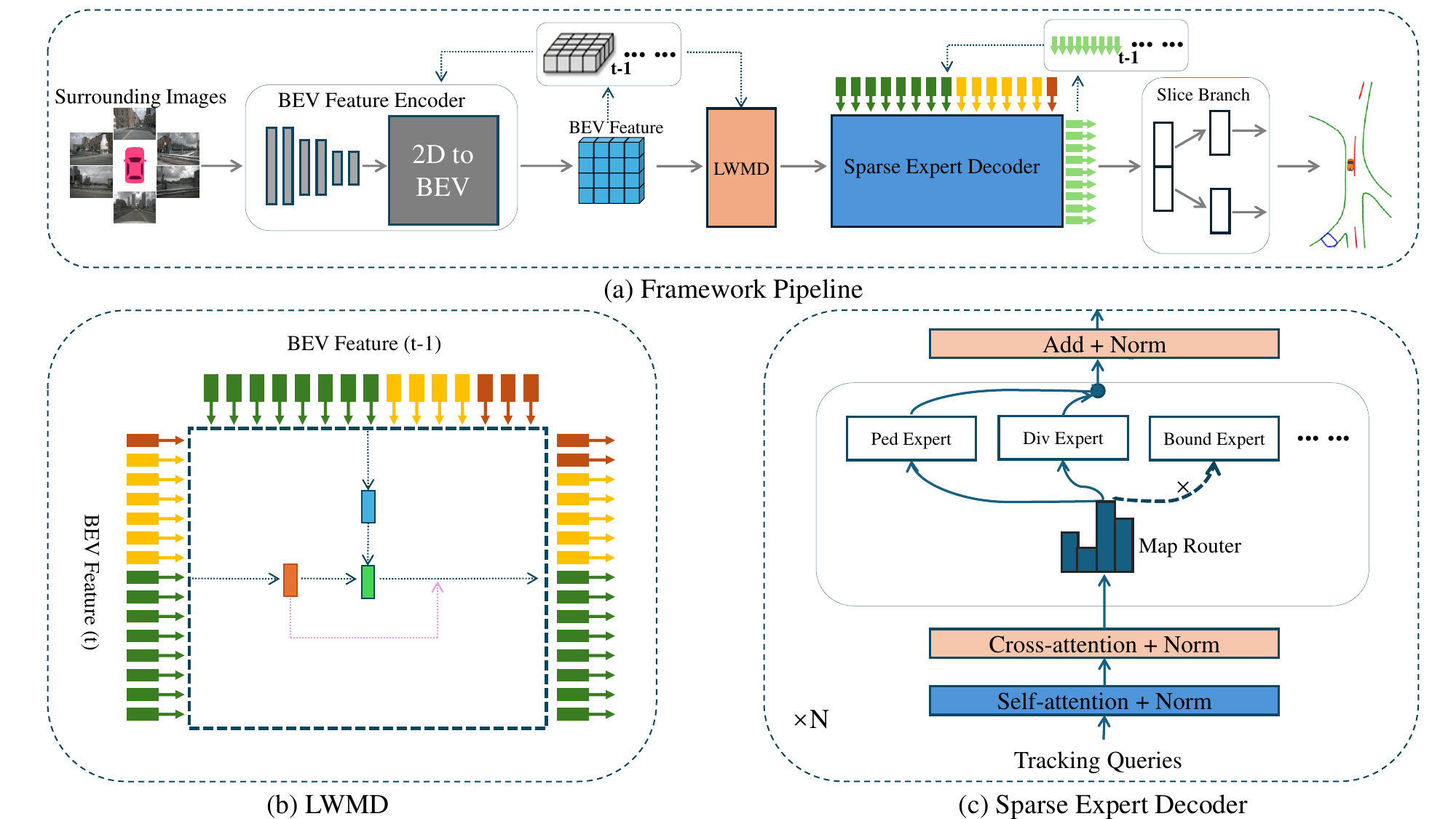} % Reduce the figure size so that it is slightly narrower than the column.
\caption{Overview of our newly introduced MapExpert: (a) The pipeline of MapExpert, consisting of our BEV feature encoder, learnable Weighted Moving Descent (LWMD), and sparse expert decoder. This pipeline processes surrounding images as input and generates vectorized map elements in an end-to-end module. (b) The detailed process of the Learnable Weighted Moving Descent, which extracts critical information from previous BEV features and enhances the representation of BEV HD map elements. (c) The structure of our unique sparse expert transformer layer is designed to effectively extract features of various map elements, such as lane dividers, pedestrian crossings, and road boundaries.}
\label{fig1}
\end{figure*}

\section{Introduction}

High-definition (HD) maps are essential for autonomous driving, conventionally constructed offline with SLAM-based methods \cite{loam, lio-sam}, along with manual annotation. However, these methods are limited by scalability issues and high maintaining costs. In recent years, bird’s-eye-view (BEV) feature extractors have introduced a novel thought \cite{lss, bevformer, Zdp7ehss}, enabling the online construction of HD Maps from BEV features. These online approaches reduce offline human efforts by predicting HD Map elements in real-time, leading to cost savings and the ability to update changes in the environment promptly.

Early researchers leveraged segmentation tasks to obtain rasterized maps based on the BEV feature maps. These methods presented each rasterized pixel as a key point and then extracted map elements and their occupancy presentation \cite{zdp1, zdp2}. With the widespread use of transformers in vision tasks, HDMapNet \cite{hdmapnet} emerged, utilizing queries to predict HD map elements. Inspired by HDMapNet, \cite{vectormapnet, mapqr, MachMap, InsMapper, HIMap, gemap, mapnext, EAN-MapNet, InstaGraM} extracted structured map information and constructed vectorized maps by sampling elements as point sets, many of these works have improved performance by designing more reasonable content queries or embedding specified positional information into the queries. Recently, some scientists introduced novel tracking-based methods to enhance the HD map prediction performance, they associate HD map elements between frames via attention queries that evolve a set of track predictions \cite{StreamMapNet, maptracker}.

However, these DETR-like methods overlook the fact that online HD map construction elements are different from traditional detection objects. Traditional detection objects, such as cars and pedestrians, are typically cube-like and relatively uniform in shape, with centralized offsets. In contrast, HD map elements are vastly different: lane dividers are normally smooth Bézier curves, road boundaries are erratic slender lines that can be jagged or closed, and pedestrian crossings are closed rectangle shapes. Fitting these diverse non-cubic map elements with a single DETR-like design is challenging. Additionally, these methods stack previous BEV features to enhance the BEV feature expression, but this can lead to current BEV features being dominated by outdated data from historical features. These factors constrain prior methods from achieving optimal performance.

In this paper, we theoretically analyze these issues and propose a novel online map construction method based on map element experts, named \textbf{MapExpert}, the architecture is illustrated in Figure \ref{fig1}. Instead of using the unified modeling methods introduced by \cite{maptr, StreamMapNet, maptracker}, which geometrically abstract different map elements into a unified representation, we employ distinct expert layers to accurately fit various map elements, such as lane dividers, pedestrian crossings, and road boundaries. We also mathematically analyze the drawbacks of stacking BEV features, and design MapExpert to not only strengthen current feature expression but also sieve and extract useful information for our decoder. To enhance the geographical position of predictions, we introduce a refined slice head branch that independently regresses map element positions and dimensions from slice tensors.

Our experiments demonstrate that MapExpert achieves state-of-the-art performance on the public nuScenes \cite{nuScenes} and Argoverse2 \cite{Argoverse2} datasets. Concretely, with the same backbone and training epochs, MapExpert reaches 76.5 mAP on nuScenes, surpassing the existing best method by 1.8 mAP. Additionally, on Argoverse2, MapExpert outperforms the existing best method by 1.4 mAP for local HD map construction, using the same backbone and training epochs. Our ablation studies illustrate that MapExpert achieves significant improvements across various and complex HD map construction scenarios. 

To summarize, the contributions of our paper are as follows:

\begin{itemize}
     \item We propose an online HD map construction method based on a novel sparse map element expert. Our approach utilizes map element routers and sparse experts specifically designed to handle map elements of varying shapes.
    \item Building on this novel modeling approach, we theoretically analyze the limitations of prevalent BEV feature temporal fusion methods and introduce an efficient temporal fusion module called Learnable Weighted  Moving Descent (LWMD). This module not only enhances the representation of current features but also filters and extracts useful information for our final BEV features.
    \item Experiments conducted on the nuScenes and Argoverse2 datasets demonstrate that our method achieves state-of-the-art performance, showing significant improvements over existing methods.

\end{itemize}

% You can remove the copyright notice and ensure that your names aren't shown by including \texttt{submission} option when loading the \texttt{aaai25} package:

% \begin{quote}\begin{scriptsize}\begin{verbatim}
% \documentclass[letterpaper]{article}
% \usepackage[submission]{aaai25}
% \end{verbatim}\end{scriptsize}\end{quote}

\section{Related Works}

\subsection{Rasterization-based Methods} 
Rasterization-based approaches construct online HD maps by extracting a rasterized Bird’s-Eye-View (BEV) representation from surrounding cameras \cite{lss, bevformer, Zdp7ehss} and then segmenting individual rasterized instances. Early methods are similar to 2D segmentation methods \cite{zdp5, zdp6, Zdp7ehss}, which usually predict the traversable area of BEV \cite{zdp4,zdp8}. BEV-LaneDet \cite{zdp3} and Persformer \cite{zdp0} treat 3D lane detection as a segmentation task based on rasterized BEV feature maps, they present each rasterized pixel as a key point to extract lane occupancy presentation. Similarly, \cite{zdp2} use simplified BEV raster representations of the surrounding scene for segmentation tasks. Furthermore, \cite{zdp1} discards positional embeddings derived from calibrated camera intrinsics and extrinsics, learning a camera-calibration-dependent mapping to predict a binary semantic segmentation mask. HDMapNet \cite{hdmapnet} also predicts semantic segmentation results on BEV features; however, unlike other rasterization-based methods, it employs complex post-processing to generate vectorized HD maps.

\subsection{Vectorization-based Methods}

Despite the use of rasterized maps, distinguishing HD map elements remains challenging. VectorMapNet \cite{vectormapnet} was the first to introduce a two-stage network for predicting sequential sampling points of HD Map elements. Unlike VectorMapNet \cite{vectormapnet}, MapTR \cite{maptr} introduces an end-to-end transformer structure that samples elements as point sets using a group of fixed permutations, this method uses hierarchical queries to extract structured map information and construct vectorized maps.
Subsequently, several studies have improved performance by designing novel hierarchical queries. These methods use scattered instance queries that share content information within the same map elements to avoid inconsistencies in the content of sampling points \cite{mapqr, MachMap, InsMapper, HIMap, gemap, mapnext, EAN-MapNet, InstaGraM}. In addition, BeMapNet \cite{bemapnet} delves a piecewise Bézier network with control point coordinates to manipulate curve shapes. 
Some papers introduce anti-disturbance methods to optimize jittery or jagged outputs \cite{admap}. Furthermore, \cite{hybrimap} enhances instance queries by adding specified positional information embedded from reference points. To alleviate the difficulty in element localization and relevant feature extraction due to the sparse and irregular detection targets, MGMap \cite{mgmap} incorporates the guidance of learned map masks with instance and point queries. Unlike other vectorization-based methods, MapVR \cite{mapvr} introduces a combined solution. 
It transforms vectorized map elements into an HD Map and then adds segmentation supervision on the rasterized HD map, experiment results present a significant improvement. Most previous approaches use a fixed number of points, which may elide essential details. PivotNet \cite{pivotnet} proposes a novel Point-to-Line mask structure to encode both subordinate and geometrical point-line priors, experiment shows a remarkably superior to others. Besides, P-MapNet \cite{p-mapnet} and MapVision \cite{MapVision} incorporate standard-definition (SD) maps and sensors to improve performance, although their application is limited due to misalignment between SD map skeletons and BEV features. Additionally, some researchers exploit the HD map performance in long-range scenarios. They propose a hierarchical sparse map construction to obtain superior performance \cite{scalablemap}. Notably, recent research introduces generative methods that combine vecterized HD map models with learned generative models for semantic map layouts, these methods obtain a better accuracy, realism, and uncertainty awareness \cite{MapPrior, PolyDiffuse}. 

\subsection{Tracking-based Methods}
Recently, researchers use transformer attention mechanism with track queries to associate tracking instances across frames \cite{trackformer, transtrack, memot, memotr}. These methods achieve significant performance improvements in tracking tasks. Inspired by these methods, StreamMapNet \cite{StreamMapNet} selects the former foregoing top k (where k is less than the maximum number of queries) queries based on confidence score as potential tracking queries, and then concatenates them with initialized queries. Experiment results demonstrate that it outperforms previous methods. Similarly, SQD-MAP \cite{sqd-map} builds upon StreamMapNet \cite{StreamMapNet} by incorporating stream query denoising \cite{dn-detr}. It feeds noised ground-truth map elements as noised queries (such as shifting, angular rotation, and scale transformation) along with learnable queries into the decoder to effectively decrease instability. MapTracker \cite{maptracker} further extends the tracking queries with memory mechanisms for better fusion. It explicitly associates tracked HD map elements from historical frames to further enhance temporal consistency. Researchers also use tracking-based HD maps as an online mapping component of end-to-end autonomous driving systems to achieve superior performance among all tasks \cite{sparsedrive}.

\section{Problem Statement}
\subsection{Inconsistent of Map Elements}   
\begin{figure}[t]
\centering
\includegraphics[width=1\columnwidth]{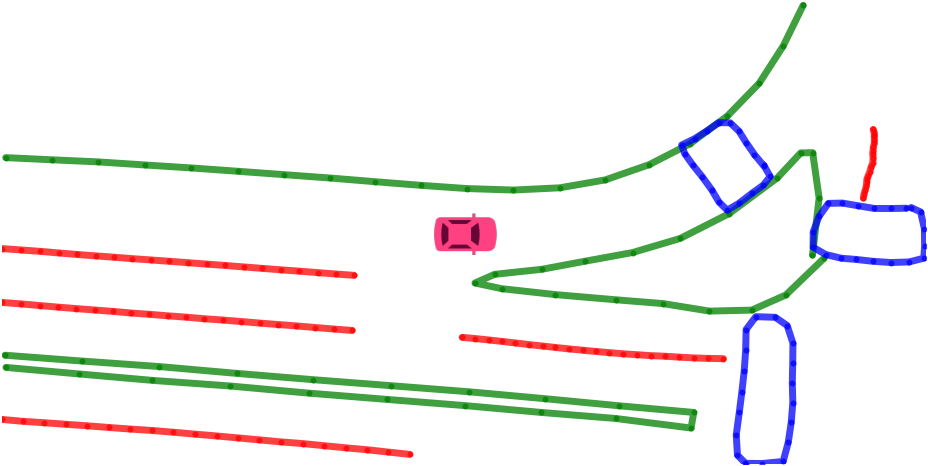}
\caption{Topology of map elements (red: lane dividers, blue: pedestrian crossings, green: road boundaries). Differ from detection objects, which are typically cube-shaped, map elements are non-cubic and can take various shapes.}
\label{fig2}
\end{figure}

Recently, most online HD map construction methods have been approached as detection tasks, involving the learning of anchors and relative offsets \cite{vectormapnet, maptr, maptracker}. While this design is undoubtedly straightforward and efficient, it overlooks a critical distinction: online HD map elements differ significantly from traditional object detection targets. Traditional detection targets typically involve cube-like objects such as vehicles, pedestrians and animals with fixed physical dimensions. These objects, which vary relatively little in shape, allow for a centralized representation of offsets across categories, simplifying the model design. It is easier to represent the features of different categories of objects using a unified model design. In contrast, as illustrated in Figure \ref{fig2}, HD map elements exhibit a wide range of geometric structures. Lane dividers are often smooth Bézier curves, road boundaries may appear as erratic, slender lines that can be jagged or closed, and pedestrian crossings are typically closed rectangles. This variability poses a significant challenge for DETR-like decoders, which are commonly used in HD map methods, as they may struggle to simultaneously capture the features of non-cubic map elements, such as Bézier curves, erratic lines, and rectangles. In addition, the diverse geometric characteristics of these map elements complicate the development of a unified geometric representation. Therefore, our analysis suggests the necessity of designing a novel approach that accurately represents various map elements without substantially increasing resource consumption.

\subsection{Dominance of Outdated Historical Data in the Current Frame}
HD map elements are static instances, which differentiates them from objects in detection tasks; theoretically, they should be easy to fuse. To achieve superior performance, researchers often incorporate historical frames to enhance the BEV feature representation. A common and effective approach is to stack the aligned historical BEV features \cite{maptr, StreamMapNet, maptracker}. For example, Maptracker \cite{maptracker} stacks four historical features. While this technique can improve performance, it also leads to an overreliance on historical frames, significantly reducing the contribution of the current frame. Additionally, it introduces substantial noise from distant historical frames. In this subsection, we will explore the nature of this issue in more details.

According to our analysis, the BEV feature stack principle reveals that each historical BEV feature contributes differently to the current stacked BEV feature used in the decoding process, as shown in Eq. 1 below:

$$ C_{t} = \frac{1}{S} F_{t} + \frac{1}{S}\sum_{n=1}^{S-1}C_{t-n} \eqno{(1)} $$

where $S(S>=2)$ denotes the number of stacked frames. $t$ indicates the frame index of $t$-th, and $F_{t}$ is the BEV feature of the $t$-th frame. $C_{t}$ is the stacked BEV feature of $t$-th frame, which is composed of the current BEV feature extracted from the current frame and $S-1$ historical stacked BEV features. Similarly, $C_{t-n}$ is the stacked BEV feature of $(t-n)$-th frame, where $n$ ranges from 1 to $S-1$. We use a solved formula to represent the stacked BEV feature $C_{t}$, as shown in Eq. 2. Furthermore, approximation methods, such as the one in Eq. 3, are used to derive an explicit approximation expression, allowing for a clearer representation of the composition of the current BEV stacked features. The forms of Eq. 2 and Eq. 3 are as follows:

$$ C_{t} = \frac{1}{S} F_{t} + \frac{1}{S}\sum_{n=1}^{t-1}W_{t-n}F_{t-n} \eqno{(2)} $$

\begin{equation}  
C_{t} \approx \left\{  
     \begin{array}{lr}  
      \frac{1}{S} F_{t} + \frac{1}{S^{2}} F_{t-1} + \frac{1}{S^{3}} F_{t-2} + \hat{R_{1}},   \quad  S = 2 &  \\  &\\ 
      \frac{1}{S} F_{t} + \frac{1}{S^{2}} F_{t-1} + \frac{S+1}{S^{3}} F_{t-2} + \hat{R_{2}}.  \quad S > 2 &    
     \end{array}  
\right.
\tag{3}
\end{equation}

where $W_{t-n}$ is the weight of $(t-n)$-th feature, which varies as $n$ increasing from 1 to $t -1$, $\hat{R_{1}}$ and $\hat{R_{2}}$ are remainders of formula. 
As indicated by Eq. 3, when the stacked number $S$ increases, the composition of the current BEV feature extracted from the current frame will decrease, which is $\frac{1}{S} F_{t}$. 
In other words, the stacked BEV feature is more likely to be dominated by historical BEV features. This is contrary to the expectation that the proportion of important information from the current BEV feature should be more reliable and substantial, which means that the current BEV feature should contribute more to current stacked BEV feature $C_{t}$. Furthermore, our analysis reveals that the proportion of HD map elements in a rasterized HD map is less than 5\%. This implies that adding more historical BEV features introduces more irrelevant information, which can even transmit noise that negatively affects HD map reconstruction. We will propose a method to address this issue effectively.

%经典三线表
\begin{table*}[h]

\centering
% \belowrulesep=0pt
% \aboverulesep=0pt
\setlength\tabcolsep{4.5pt}
\renewcommand{\arraystretch}{1}
% \normalsize %字体
%\resizebox{140mm}{35mm}{
% \tiny{

\begin{tabular}{c|cc|cccc|cccc}%四个c代表有四列且内容居中
\hline%第一道横线
\multirow{2}*{\textbf{Methods}} &\multirow{2}*{\textbf{Backbone}} &\multirow{2}*{\textbf{Epoch}}  &\multicolumn{1}{c}{\textbf{AP\tiny{ped}}} &\multicolumn{1}{c}{\textbf{AP\tiny{div}}} &\multicolumn{1}{c}{\textbf{AP\tiny{bound}}}  &\multicolumn{1}{c|}{\textbf{{mAP}}} &\multicolumn{1}{c}{\textbf{{AP\tiny{ped}}}} &\multicolumn{1}{c}{\textbf{{AP\tiny{div}}}} &\multicolumn{1}{c}{\textbf{{AP\tiny{bound}}}}  &\multicolumn{1}{c}{\textbf{{mAP}}} \\%跨两列;内容居中;跨列内容为Resultsummary通过\textbf{}与\underline{}命令分别对内容加粗、加下划线
& & & \multicolumn{4}{c|}{Hard: \{0.2, 0.5, 1.0\}m} & \multicolumn{4}{c}{Easy: \{0.5, 1.0, 1.5\}m} \\
\hline%第二道横线
HDMapNet\cite{hdmapnet}&EB0&120 &\---&\---&\---&\---&14.4&21.7&33.0&23.0\\
\hline%第二道横线
VectorMapNet\cite{vectormapnet}	&R50&110	&20.6&32.4&	24.3&	25.7&	42.5&	51.4	&44.1&	46.0\\
\hline%第二道横线
MapTR\cite{maptr}	&R50	&110 &31.4&	40.5&35.5&35.8	&56.2	&59.8 &60.1	& 58.7 \\
\hline%第二道横线
MapTRv2\cite{maptrv2}	&R50	&110 &43.6	&49&	43.7&	45.4&	68.1&68.3&	69.7	&68.7\\
\hline%第二道横线
MapQR\cite{mapqr}	&R50	&110	&46.2	&57.3	&48.1	&50.5	&67.1&70.4	&71.2&	69.6\\
\hline%第二道横线
StreamMapNet\cite{StreamMapNet}	&R50	&110	&44.4&	60.5	&48.6	&51.2	&68	&71.2	&69.2	&69.5\\
\hline%第二道横线
MapTracker\cite{maptracker}	&R50	&100	&52.3	&62.5	&59.6&	58.13	&77.3	&72.4	&74.2	&74.7\\
\hline%第二道横线
Proposed	&R50	&100	&\textbf{55.7}&\textbf{64.2}	&\textbf{61.3}	&\textbf{60.4}	&\textbf{79.4}	&\textbf{73.9}	&\textbf{76.2}	&\textbf{76.5} 	\\
% \midrule%第二道横线

\hline%第四道横线
\end{tabular}
\caption{Performance comparison of various methods on the original nuScenes split at both 50m and 30m perception ranges. Specifically, results are evaluated on \{0.2 m, 0.5 m, 1.0 m\} and \{0.5 m, 1.0 m, 1.5 m\} thresholds. MapExpert notably outperforms other methods across all categories. The best results for same settings (i.e., backbone and epoch) are highlighted in bold.}
% }
%}
\label{table1}
\end{table*}

\begin{table}[h]

\centering
% \belowrulesep=0pt
% \aboverulesep=0pt
\setlength\tabcolsep{1pt}
\renewcommand{\arraystretch}{1}
% \tiny %字体
% \small
%\resizebox{140mm}{35mm}{
% \resizebox{0.47\textwidth}{10mm}{
% \tiny{

\begin{tabular}{c|cccc}%四个c代表有四列且内容居中
\hline%第一道横线
\multirow{1}*{\textbf{Methods}}  &\multicolumn{1}{c}{\textbf{AP\tiny{ped}}} &\multicolumn{1}{c}{\textbf{AP\tiny{div}}} &\multicolumn{1}{c}{\textbf{AP\tiny{bound}}}  &\multicolumn{1}{c}{\textbf{{mAP}}} \\%跨两列;内容居中;跨列内容为Resultsummary通过\textbf{}与\underline{}命令分别对内容加粗、加下划线
% & & & \multicolumn{4}{c}{} \\%{Easy: \{0.5, 1.0, 1.5\}m}  \\
\hline%第二道横线
StreamMapNet\cite{StreamMapNet}	&31.6 &	28.1 &	40.7 &	33.5
\\
\hline%第二道横线
MapTracker\cite{maptracker}	&45.9 &30.0 	&45.1 &	40.3	\\
\hline%第二道横线
Proposed	&\textbf{46.7}	&\textbf{34.1}	&45.1	&\textbf{42.0}	\\
% \midrule%第二道横线

\hline%第四道横线
\end{tabular}
% }
% }
\caption{Comparison with SOTA Methods on the new nuScenes split validation set. All the experiments are based on ResNet50 backbone. Results are evaluated on \{0.5 m, 1.0 m, 1.5 m\} thresholds.}
\label{table2}
\end{table}

\section{Methods}
\subsection{Architecture Overview}

The overall model architecture is illustrated in Figure \ref{fig1}, and is streamlined into three components as follows: 

\textbf{BEV Encoder:} 
Given surrounding images $P \in \mathbb{R}^{N \times 3\times H\times W}$ from multi-camera setups, we extract multi-scale image features using ResNet \cite{resnet} and FPN \cite{fpn}. These multi-scale features are then fed into a 2D-to-BEV transformer encoder \cite{bevformer, maptracker}. Our BEV queries are initialized with the previously aligned BEV feature, to obtain the BEV feature map.

\textbf{Learnable Weighted Moving Descent:}
Most existing methods concatenate the previously aligned hidden states of BEV features with the current BEV feature to enhance the expression of BEV HD map elements. However, as mentioned earlier, stacked BEV features are more likely to be dominated by historical BEV features, despite the current BEV feature's potential for greater contribution. In this method, we mathematically analyze this issue and propose a novel strategy named Learnable Weighted  Moving Descent (LWMD) to fuse the BEV features reasonably.
 
\textbf{Sparse Expert Decoder:}
We utilize a hierarchical tracking query scheme to explicitly fuse historical tracking features and extract map elements with our sparse expert transformer layer. Specifically, we initialize the tracking query as $TQ(t) = [TQ_{element}(t-1), TQ_{init}]$. $TQ(t)$ is the final current tracking query that will be fed into the sparse expert transformer layer. $TQ_{element}(t-1)$ denotes the aligned previous decoder output, $TQ_{init}$ presents the initialized empty query, which can be used to pad the $TQ(t)$ to the designed size. This tracking strategy is borrowed from \cite{memotr, maptracker}. Subsequently, we send the $TQ(t)$ and final BEV feature into our sparse expert transformer layer, where the map element experts of our sparse expert transformer layer will extract excellent map element characteristics. Details of the sparse expert transformer layer will be provided later.

%经典三线表
\begin{table*}[h]

\centering
% \belowrulesep=0pt
% \aboverulesep=0pt
\setlength\tabcolsep{14.5pt}
\renewcommand{\arraystretch}{1}
% \tiny %字体
%\resizebox{140mm}{35mm}{
% \resizebox{0.47\textwidth}{10mm}{
% \resizebox{0.47\textwidth}{10mm}{
% \tiny{

\begin{tabular}{c|cc|cccc}%四个c代表有四列且内容居中
\hline%第一道横线
\multirow{1}*{\textbf{Methods}} &\multirow{1}*{\textbf{Backbone}} &\multirow{1}*{\textbf{Epoch}}  &\multicolumn{1}{c}{\textbf{AP\tiny{ped}}} &\multicolumn{1}{c}{\textbf{AP\tiny{div}}} &\multicolumn{1}{c}{\textbf{AP\tiny{bound}}}  &\multicolumn{1}{c}{\textbf{{mAP}}} \\%跨两列;内容居中;跨列内容为Resultsummary通过\textbf{}与\underline{}命令分别对内容加粗、加下划线
% & & & \multicolumn{4}{c}{} \\%{Easy: \{0.5, 1.0, 1.5\}m}  \\
\hline%第二道横线
HDMapNet\cite{hdmapnet}&EB0&120 &13.1  &5.7 & 37.6 	&18.8 \\
\hline%第二道横线
VectorMapNet\cite{vectormapnet}	&R50&110	&36.5  &35.0 	&36.2&35.8\\
\hline%第二道横线
MapTR\cite{maptr}	&R50	&110 & 55.4&	58.7&59.1&57.8 \\
\hline%第二道横线
MapTRv2\cite{maptrv2}	&R50	&110 &60.7&68.9&64.5&64.7\\
\hline%第二道横线
MapQR\cite{mapqr}	&R50	&110	&71.2 &	60.1 &66.2&65.9\\
\hline%第二道横线
StreamMapNet\cite{StreamMapNet}	&R50	&110&64.9	&60.2	&64.9&63.3\\
\hline%第二道横线
MapTracker\cite{maptracker}	&R50	&100	&74.5&66.4&73.4	&71.4\\
\hline%第二道横线
Proposed	&R50	&100	&\textbf{76.4}	&\textbf{66.9}	&\textbf{75.1}	&\textbf{72.8}		\\
% \midrule%第二道横线

\hline%第四道横线
\end{tabular}
% }
% }
\caption{Performance comparison with baseline methods on the original Argoverse2 split at 30 m perception ranges. MapExpert outperforms existing state-of-the-art methods. Results are evaluated using thresholds of \{0.5 m, 1.0 m, 1.5 m\}. }
\label{table3}
\end{table*}

%经典三线表
\begin{table}[h]

\centering
% \belowrulesep=0pt
% \aboverulesep=0pt
\setlength\tabcolsep{1pt}
\renewcommand{\arraystretch}{1}
% \tiny %字体
%\resizebox{140mm}{35mm}{
% \resizebox{0.47\textwidth}{10mm}{
% \resizebox{0.47\textwidth}{10mm}{
% \tiny{

\begin{tabular}{c|cccc}%四个c代表有四列且内容居中
\hline%第一道横线
\multirow{1}*{\textbf{Methods}}  &\multicolumn{1}{c}{\textbf{AP\tiny{ped}}} &\multicolumn{1}{c}{\textbf{AP\tiny{div}}} &\multicolumn{1}{c}{\textbf{AP\tiny{bound}}}  &\multicolumn{1}{c}{\textbf{{mAP}}} \\%跨两列;内容居中;跨列内容为Resultsummary通过\textbf{}与\underline{}命令分别对内容加粗、加下划线
% & & & \multicolumn{4}{c}{} \\%{Easy: \{0.5, 1.0, 1.5\}m}  \\
\hline%第二道横线
StreamMapNet\cite{StreamMapNet}	&61.8 &68.2 &63.2 &	64.4\\
\hline%第二道横线
MapTracker\cite{maptracker}		&70.0 &75.1 &68.9 &71.3\\
\hline%第二道横线
Proposed	&\textbf{71.2}	&\textbf{75.6}	& 68.7	&\textbf{71.8}	\\
% \midrule%第二道横线

\hline%第四道横线
\end{tabular}
% }
% }
\caption{Comparison with state-of-the-art method on new Argoverse2 split,  following the evaluation criteria used in Table \ref{table3}. Results are evaluated on \{0.5 m, 1.0 m, 1.5 m\} thresholds. }
\label{table4}
\end{table}

\subsection{Sparse Expert Transformer Layer}

As analyzed in the "Inconsistent of Map Elements" subsection, most online HD Map construction methods are detection tasks, which involve learning anchors and relative offsets. This design is undoubtedly simple and efficient. However, traditional detection objects, such as cube-like vehicles and animals with fixed physical dimensions, are significantly different in shape from map elements like bézier curves, erratic slender lines, and rectangles. Therefore, using a detection-based design to fit these map elements is inappropriate. 

Inspired by Mixture of Experts \cite{moe}, which selects different parameters for each incoming example, we have designed a novel decoder layer named the sparse expert transformer layer. This layer mainly consists of self-attention, cross-attention, and sparse map element experts. A brief overview of this layer is illustrated in Figure \ref{fig1}. First, self-attention takes tracking queries as inputs to obtain a representation, then extracts map features with cross-attention. The output from cross-attention is fed into our sparse map element expert, which is composed by rooters and map element experts. Each expert is responsible for specific map elements (lane dividers, pedestrian crossings, and road boundaries). Concretely, our map expert router primarily routes the representation from the previous module to the best-determined top-K experts selected from a set $\{E_{i}(x)\}_{i=0}^{N-1}$, where $N$ is the total number of experts, and $E_{i}$ is the $i$-th expert. Briefly, we normally do not compute the outputs of experts whose routes are zero, this sparse route could limit the computation costs. Our routers are implemented by normalizing via a softmax distribution over the top-K logits. As illustrated below:
$$ R(x) = SoftMax(TopK(x \cdot W_{r})) \eqno{(4)} $$
where $x$ is the map element feature, $R(x)$ denotes the output of the map expert router, $R(x)= ri$ if $ri$ is among the top-K coordinates of logits, and $R(x)= -\infty$ otherwise. The router variable $W_{r}$ produces logits $x \cdot W_{r}$. The value K of top-K is a hyper-parameter that modulates the amounts of experts used to process map elements. This design has a notable success in computational efficiency, which means that even if we increase $N$ while keeping $K$ fixed, the model’s parameter count increases while the computational cost remains constant. This motivates a distinction between the model’s total parameter count and the number of parameters used for processing an individual active parameter count, also known as sparse expression.	Our map element feature $x$ meant to be processed by specific experts, is routed to the corresponding expert for processing. The expert’s output is then returned to the original query position. As shown in Figure \ref{fig1}, we design three types of experts: the lane divider experts, the pedestrian crossing experts, and the road boundary experts. These experts are intended to extract different types of map element features, such as bézier curves, erratic slender lines, and rectangles. We use the expert router to select top-K experts from a set $\{E_{i}(x)\}_{i=0}^{N-1}$ expert networks, the simplified expression of the sparse map element expert is given by:
$$ y = \sum_{i=0}^{N-1}R_{i}(x) E_{i}(x) \eqno{(5)} $$
here, $y$ is the output, $x$ is the map element feature, $N$ is the total experts count, $R_{i}(x)$ is the $i$-th router output, and $E_{i}(x)$ is the $i$-th sparse map element expert. Concretely, we use the SwiGLU as the expert $E_{i}(x)$, which means that each map element feature $x$ is routed to K SwiGLU blocks with different sets of weights. The output $y$ for an input token $x$ is represented as:
$$ y = \sum_{i=0}^{N-1} SoftMax(TopK(x \cdot W_{r}))_{i} \cdot SwinGLU_{i}(x) \eqno{(6)} $$

Note that this final formulation introduces challenges in load balancing, which will be analyzed in the auxiliary loss subsection.

\subsection{Learnable Weighted Moving Descent} 
As illustrated in Eq.3, existing methods usually concatenate historical BEV features, which can lead to the issue analyzed in the second part of the problem statement: an over-reliance on historical BEV features can disproportionately influence the current BEV feature. Therefore, this paper proposes a novel module called Learnable Weighted Moving Descent (LWMD), which integrates historical BEV features into the current BEV features without increasing device memory. Our LWMD uses a single previous BEV feature to achieve superior performance compared to stacking multiple BEV frames. Our approach is a learnable method that automatically adjusts the fusion between features. The formula for LWMD is as follows:
$$ C_{t} = \beta F_{t} + f_{t}(F_{t},C_{t-1}) \eqno{(7)} $$
% $$ C_{t} = \beta F_{t} + f_{t}(F_{t}, \beta F_{t-1} + f_{t-1}(F_{t-1}, \beta F_{t-2} + f_{t-2}(F_{t-2},C_{t-3})) $$
here, the final fused result $C_{t}$ consists of two components: the current BEV feature $F_{t}$ and the map element information extracted from the current BEV feature and the last fused BEV feature via $f_{t}$, $f_{t}$ is a neural network received the current BEV feature $F_{t}$ and the previous fused BEV feature $C_{t-1}$ as inputs. $\beta$ is a learnable parameter. In our formula, the proportion of current BEV features is learnable, thereby mitigating the issue of historical data dominance as highlighted in the problem statements. To sum up, our approach, which fuses two frames of features, achieves performance comparable to or better than that of stacking multiple BEV frames.

\subsection{Auxiliary Expert Balance Loss} 
Without a balance strategy, experts may encounter an inhomogeneous situation. For example, one expert might be trained to handle all three map elements, while others may be skipped forever. To encourage a balanced load across different experts, we add an auxiliary loss named the auxiliary expert balance loss for the sparse expert transformer layer. This additional loss encourages each expert to be of equal importance. Differing from \cite{moe, GShard}, we distribute the workload evenly with a simplified design. Given N experts indexed by i = 0 to $N-1$ and $T$ tokens, the auxiliary expert balance loss is calculated as the scaled dot product between parameters $f$ and $P$.
$$ L_{expert-balance} = \alpha \cdot N \cdot \sum_{i=0}^{N-1}f_{i} \cdot P_{i} \eqno{(8)} $$

$$ f_{i} = \frac{1}{T} OneHot(TopK(\frac{e^{p_{i}(x)}}{\sum_{j=0}^{N-1}e^{p_{j}(x)}} )) \eqno{(9)} $$

$$ P_{i} = \frac{1}{T} p_{i}(x) \eqno{(10)} $$
here, $f_{i}$ is the percentage of inputs routed to each expert, $\alpha$ is a hyper-parameter, and $P_{i}$ is the fraction of the router probability allocated to each corresponding expert, $p_{i}$ is the probability of routing token $x$ to expert $i$.

The auxiliary expert balance loss, as described in Eq.8, ensures uniform routing for three map elements. This loss function is differentiable thanks to the $P_{i}$. Finally, we add an auxiliary balance loss to the total loss during training.

\begin{table*}[h!]

\centering
% \belowrulesep=0pt
% \aboverulesep=0pt
\setlength\tabcolsep{8pt}
\renewcommand{\arraystretch}{1}
% \tiny %字体
%\resizebox{140mm}{35mm}{
% \resizebox{0.47\textwidth}{10mm}{
% \resizebox{0.47\textwidth}{10mm}{
% \tiny{

\begin{tabular}{c|cccc|cccc}%四个c代表有四列且内容居中
\hline%第一道横线 $W_{t-n}$
\multirow{1}*{\textbf{Index}} &\multirow{1}*{\textbf{+ Expert}} &\multirow{1}*{\textbf{+ Expert Balance loss}} &\multirow{1}*{\textbf{+ LWMD}} &\multirow{1}*{\textbf{+ Slice Branch}}&\multicolumn{1}{c}{\textbf{AP\tiny{ped}}} &\multicolumn{1}{c}{\textbf{AP\tiny{div}}} &\multicolumn{1}{c}{\textbf{AP\tiny{bound}}}  &\multicolumn{1}{c}{\textbf{mAP}} \\%跨两列;内容居中;跨列内容为Resultsummary通过\textbf{}与\underline{}命令分别对内容加粗、加下划线
% & & & \multicolumn{4}{c}{} \\%{Easy: \{0.5, 1.0, 1.5\}m}  \\
\hline%第二道横线
0 & & &  & & 77.6	&71.2&	73.0	&73.9 \\
\hline%第二道横线
1 &\CheckmarkBold& &  & & 78.3	&72.7	&73.9&	75.0 \\
\hline%第二道横线
2&\CheckmarkBold & \CheckmarkBold	&  & &79.0	&73.5	&76.1&76.2\\
\hline%第二道横线
3&	\CheckmarkBold&\CheckmarkBold & \CheckmarkBold&	&79.5&	73.4&76.4	&76.4 \\
\hline%第二道横线
4&\CheckmarkBold&\CheckmarkBold &\CheckmarkBold &\CheckmarkBold &79.4	&73.9	&76.2&	\textbf{76.5}\\

\hline%第四道横线
\end{tabular}
% }
% }
\caption{Ablation studies on the key design elements of MapExpert, evaluated on the origin nuScenes split dataset. Results show that each modification contributes to the performance gain.}
\label{table5}
\end{table*}
\section{Experiments}
\subsection{Experimental Settings}
% Our experimental settings are as follows:

\subsubsection{Datasets.} We evaluate our MapExpert on the nuScenes \cite{nuScenes} and Argoverse2 datasets \cite{Argoverse2}. The nuScenes is a comprehensive, synthetically generated autonomous driving dataset, consisting of 1000 scenes with annotations at 2 Hz. Each frame contains data from six synchronized surrounding cameras. We use the 2D vectorized map elements provided by nuScenes as the ground truth. Argoverse2 is another large-scale benchmark with approximately 108,000 frames, each offering images from seven surrounding cameras. Differing with nuScenes, Argoverse2 provides 3D vectorized map elements as ground truth. 
\subsubsection{Implementation Details.}
We follow the majority of the settings from MapTracker \cite{maptracker}, which uses ResNet50 \cite{resnet} and BEVFormer \cite{bevformer} for BEV feature extraction, we then conduct with our decoder to obtain a refined prediction result. We perform our experiments on six A800 GPUs. MapExpert has notable successes in local HD map construction, however, it suffers from training instabilities. To address this, we apply several training techniques. First, we incorporate an additional segmentation loss to facilitate convergence, as the vectorized loss may cause divergence during training. Second, we use a large batch size to prevent getting trapped in local optima or experiencing complete divergence.

\subsection{Comparisons with State-of-the-art Methods}

We implemented our method based on the approaches described in \cite{maptr, StreamMapNet, maptracker}. We also adopted the dataset split strategy of StreamMapNet, and evaluated our method with both the original and new split strategies. Table \ref{table1} and Table \ref{table3} follow the original split strategy, Table \ref{table2} and Table \ref{table4} follow the new split strategy of StreamMapNet. These split strategies only differ in the division between the training set and the validation set. For a fair comparison, we evaluated our method and other methods using distinct thresholds: \{1.0 m, 1.5 m, 2.0 m\} for the 50 m range and \{0.5 m, 1.0 m, 1.5 m\} for the 30 m range. 

\subsubsection{Performance on nuScenes.}
We provide a comparison of MapExpert’s performance against existing methods to ensure a comprehensive analysis. As illustrated in Table \ref{table1}, our method achieves a better mAP with the original dataset split ground truth. Concretely, MapExpert significantly outperforms the competing methods by more than 2.4\% in mAP scores with the original dataset split. Note that, MapExpert achieves 79.4 AP of pedestrian crossings, which is 2.1 mAP higher than the result of the previous best-performing method. We further compare our method with existing methods using the new NuScenes split dataset. As shown in Table \ref{table2}, our approach outperforms MapTracker by a notable margin (with improvements of +1.7 mAP overall, +4.1 AP in lane dividers, +0.9 AP in pedestrian crossings).

\subsubsection{Performance on Argoverse2.}
We also evaluated our method on Argoverse2 datasets. Table \ref{table3} shows the comparison on the original Argoverse2 split. Our method achieves superior performance over previous best-performing methods across all map elements, with 1.4 mAP higher than MapTracker and 9.5 mAP higher than StreamMapNet. Based on geographically non-overlapping splits proposed by StreamMapNet, Table \ref{table4} reveals performance on the new Argoverse2 split. The experiments on the new Argoverse2 dataset split demonstrate the superior construction ability of MapExpert in local HD map construction. We achieve 71.8 mAP, which is 0.5 mAP higher than MapTracker.

\subsection{Ablation Studies}

\subsubsection{Key Components Design.} 
We conducted several ablation studies on the original nuScenes split to confirm the necessity of the proposed modules. Initially, we integrated our modules into the baseline of MapTracker, and the influence of each component is presented in Table \ref{table5}. It is evident that all design elements in our MapExpert contribute to performance improvements, thereby validating their necessity. The index 0 is baseline. The second variant incorporates map expert components into the decoder modules, which induces approximately a 0.4\% performance increase over the baseline. The third variant includes both the experts and the auxiliary expert balance loss components, this variant surpasses the baseline by 1.5 mAP, demonstrating the effectiveness of our mechanism. The fourth variant incorporates our LWMD after the BEV encoder. This design enables the model to extract useful information from historical features without the need to stack BEV features, resulting in a 0.2 mAP improvement. Additionally, we introduced a slice branch to predict map elements. Although this modification only results in a 0.1 mAP improvement.

Other Ablations are detailed in Appendix.

\section{Conclusion}
MapExpert is a simple and efficient online HD map construction method that introduces a sparse map element expert transformer architecture and Learnable Weighted Moving Descent (LWMD) strategy to model road structure topology based on tracking-based methods. Extensive experiments demonstrate that it significantly outperforms existing methods on nuScenes and Agroverse2 datasets.

% \section{Acknowledgments}

\bibliography{aaai24}

\clearpage

% \title{MapExpert: Online HD Map Construction with Simple and Efficient\\Sparse Map Element Expert}
% \section{\textbf{Supplementary Material}}
%经典三线表

% \section{Appendix}
% \appendix
% \subsection{More Ablation}
\textbf{Limitations:} This paper identifies two limitations. First, to improve the performance, we select 2 experts from 8, leaving the remaining 6 experts inactive during module inference; however, they still consume memory. Second, with our experts and LWMD design, our model is challenging to train, it normally needs a large batch size to achieve optimal convergence results.

\subsubsection{Other Ablations.}
We also conducted several quantitative ablation studies to demonstrate the effectiveness of the map expert design, which are detailed in \textbf{Appendix A} and \textbf{Appendix B}. Furthermore, we performed additional qualitative comparisons, for comprehensive details on all qualitative experiments, please refer to \textbf{Appendix C}.

\section{Appendix A}

\begin{figure}[h]
\centering
\includegraphics[width=1.0\columnwidth]{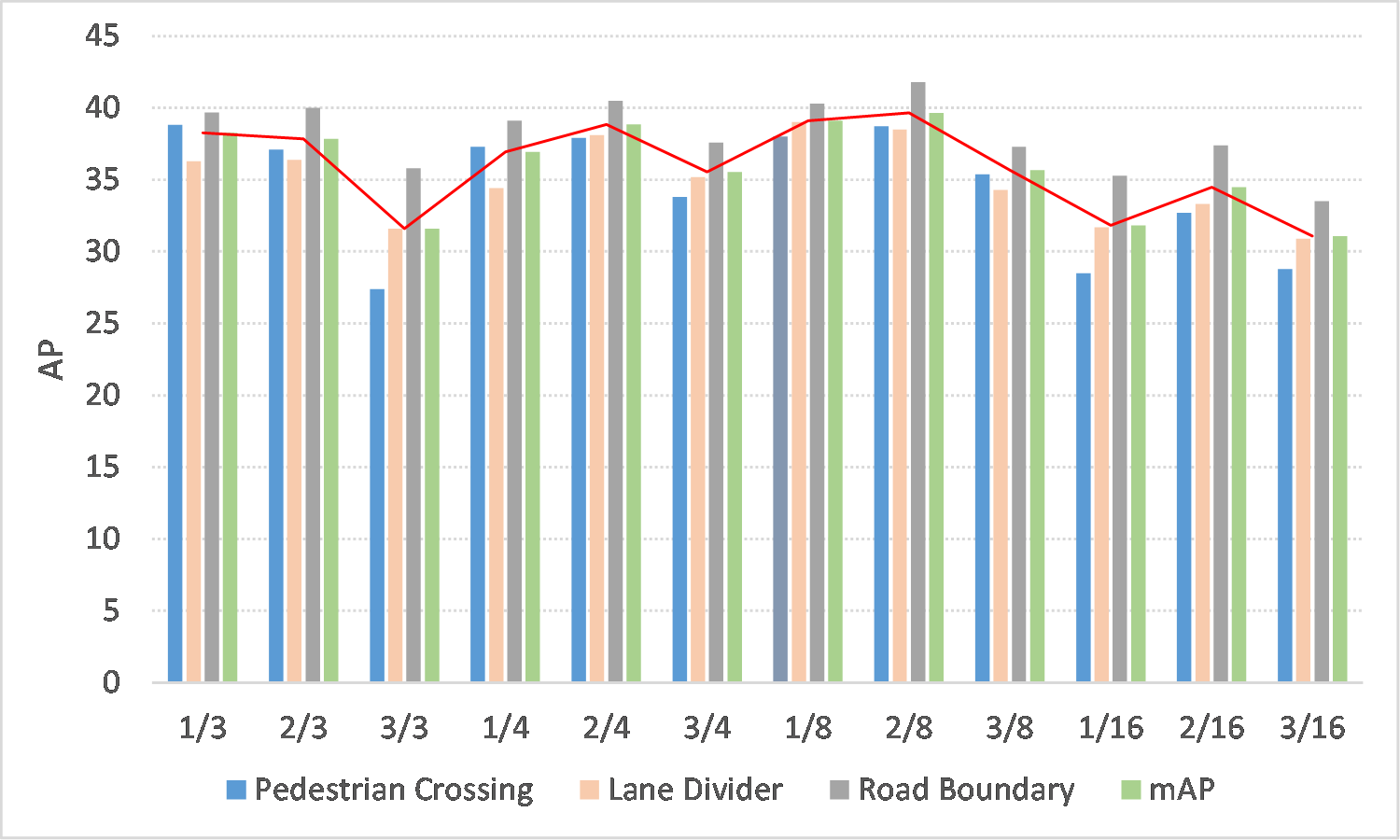} % Reduce the figure size so that it is slightly narrower than the column. Don't use precise values for figure width.This setup will avoid overfull boxes.
\caption{Ablation studies on the expert quota, evaluated on the original nuScenes split dataset.}
\label{fig3}
\end{figure}

\begin{figure*}[h]
\centering
\includegraphics[width=0.95\textwidth]{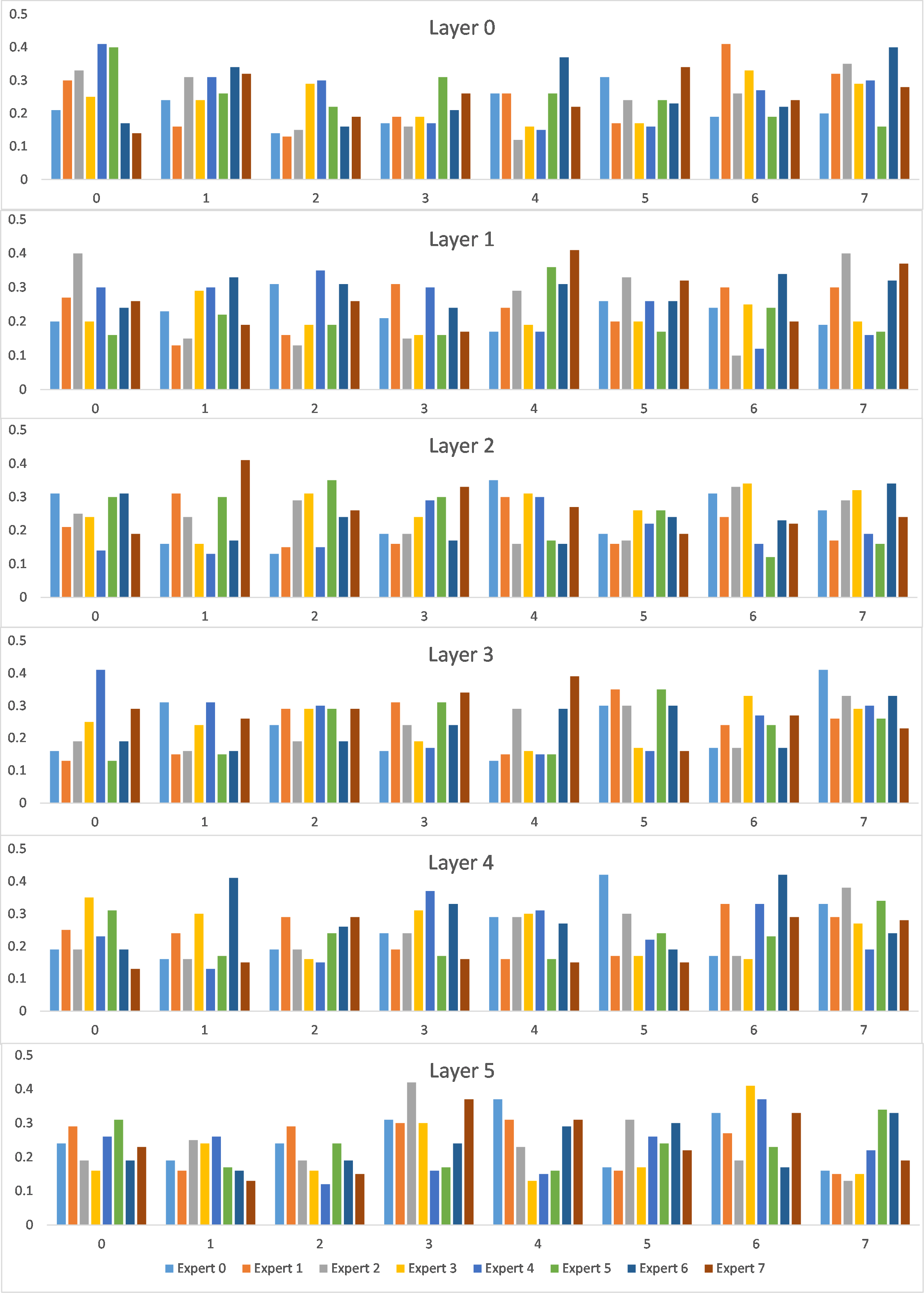} % Reduce the figure size so that it is slightly narrower than the column. Don't use precise values for figure width.This setup will avoid overfull boxes.
\caption{Proportion of queries assigned to each expert across different frames from nuScenes dataset, separated by whether the expert was selected as first or second choice, or either.}
\label{fig4}
\end{figure*}

\subsubsection{Expert Quota.}
To further investigate the impact of our sparse map element expert, we performed another ablation study focusing on the expert quota. As shown in Figure \ref{fig3}, we use different numbers of experts per layer, resulting in a linear increase in the number of weight parameters. Here we conducted our ablation studies with a brief training duration of 20 epochs. We consider 12 models with identical depths, with an increasing number of experts per layer: 3, 4, 8, and 16. As we increase the number of experts per layer from 3 to 16, we notice that the performance peaked with 8 experts per layer and 2 routers. Although this increase in the number of experts led to significant quality improvements, it also indicated the potential for increased memory consumption. To balance performance with training and inference efficiency, we  selected 8 experts and 2 routers for our experimental setup.

\subsubsection{Router Analysis.}
We aimed to determine whether each expert specialized in specific domains (e.g. lane dividers, pedestrian crossings, road boundaries). To investigate this, we analyzed the distribution of expert selection on nuScenes datasets. Results are presented in Figure \ref{fig4}, for each layer (where layers 0 and 5 represent the first and the last sparse map element expert transformer layer in the sparse expert decoder). Surprisingly, across all layers, the distribution of expert assignments is marginally different. This indicates that the nature of map elements is effectively classified by the routers. Thanks to our sparse map element expert design, the same map elements are often routed through the same expert, even when involving multiple tokens, the proportion of repeated assignments is significantly higher than random routing for the expert layers.

% \section{\textbf{Supplementary Material1}}
% \subsection{More Ablation}
\subsubsection{More Sparse Expert Transformer Layers.}
In Table \ref{table6}, more detailed experimental results are provided, including the average precision (AP) for lane dividers, pedestrian crossings, and road boundaries respectively. The results indicate that increasing the depth of the sparse expert transformer layers leads to improved performance.
%经典三线表
\begin{table}[h]

\centering
% \belowrulesep=0pt
% \aboverulesep=0pt
\setlength\tabcolsep{1pt}
\renewcommand{\arraystretch}{1}
% \tiny %字体
\small %字体
%\resizebox{140mm}{35mm}{
% \resizebox{0.47\textwidth}{10mm}{
% \resizebox{0.47\textwidth}{10mm}{
% \tiny{

\begin{tabular}{c|cccc}%四个c代表有四列且内容居中
\hline%第一道横线
\multirow{1}*{\textbf{Sparse Expert Transformer Layers}}  &\multicolumn{1}{c}{\textbf{AP\tiny{ped}}} &\multicolumn{1}{c}{\textbf{AP\tiny{div}}} &\multicolumn{1}{c}{\textbf{AP\tiny{bound}}}  &\multicolumn{1}{c}{\textbf{{mAP}}} \\%跨两列;内容居中;跨列内容为Resultsummary通过\textbf{}与\underline{}命令分别对内容加粗、加下划线
% & & & \multicolumn{4}{c}{} \\%{Easy: \{0.5, 1.0, 1.5\}m}  \\
\hline%第二道横线
1	&35.4	&38.8&	40.1&	38.1\\
\hline%第二道横线
2		&38.7&	38.5&	41.8&	39.7\\
\hline%第二道横线
3	&38.3&	39.7&	41.6&	39.9	\\
\hline%第二道横线
4	&39.3	&39.7&	43.2&	40.7	\\
% \midrule%第二道横线

\hline%第四道横线
\end{tabular}
% }
% }
\caption{The average precision (AP) on the original nuScenes split validation set with sparse expert transformer layers increases. All results are conducted with 20 training epochs.}
\label{table6}
\end{table}

\section{Appendix B}

\subsubsection{Modality Evaluation.}
We also trained our module using additional sensors, such as LiDAR. As shown in Table \ref{table7}, with the schedule of only 20 epochs, multi-modality MapExpert significantly outperforms our base module result by 5.6 mAP.

%经典三线表
\begin{table}[h]

\centering
% \belowrulesep=0pt
% \aboverulesep=0pt
\setlength\tabcolsep{7.5pt}
\renewcommand{\arraystretch}{1}
% \tiny %字体
\small %字体
%\resizebox{140mm}{35mm}{
% \resizebox{0.47\textwidth}{10mm}{
% \resizebox{0.47\textwidth}{10mm}{
% \tiny{

\begin{tabular}{c|cccc}%四个c代表有四列且内容居中
\hline%第一道横线
\multirow{1}*{\textbf{Modality}}  &\multicolumn{1}{c}{\textbf{AP\tiny{ped}}} &\multicolumn{1}{c}{\textbf{AP\tiny{div}}} &\multicolumn{1}{c}{\textbf{AP\tiny{bound}}}  &\multicolumn{1}{c}{\textbf{{mAP}}} \\%跨两列;内容居中;跨列内容为Resultsummary通过\textbf{}与\underline{}命令分别对内容加粗、加下划线
% & & & \multicolumn{4}{c}{} \\%{Easy: \{0.5, 1.0, 1.5\}m}  \\
\hline%第二道横线
Camera	&38.7&	38.5&	41.8&	39.7\\
\hline%第二道横线
LiDAR		&41.2	&39.9&	45.4&	42.2\\
\hline%第二道横线
Camera \& LiDAR	&43.5	&44.3	&48.2&	45.3	\\

\hline%第四道横线
\end{tabular}
% }
% }
\caption{Ablations about the modality. All results are conducted with 20 training epochs.}
\label{table7}
\end{table}

\subsubsection{Robustness to the Router Distribution Deviation.} 

To improve the robustness of our expert router distribution, we introduced randomly generated jitter noise to the router inputs, with the noise following a normal distribution. Results show that the standard deviation has a significant impact on performance. As illustrated in Table \ref{table8}, when the standard deviation is small, MapExpert still maintains comparable performance.

\begin{table}[h]

\centering
% \belowrulesep=0pt
% \aboverulesep=0pt
\setlength\tabcolsep{2.5pt}
\renewcommand{\arraystretch}{1}
% \normalsize %字体
% \resizebox{90mm}{15mm}{
% \tiny{

\begin{tabular}{c|cccccccc}%四个c代表有四列且内容居中
\hline%第一道横线
\multirow{2}*{\textbf{Methods}}  
& \multicolumn{8}{c}{\textbf{Jitter Ratio}}\\

&\multicolumn{1}{c}{\textbf{0} }
&\multicolumn{1}{c}{\textbf{0.001} }
&\multicolumn{1}{c}{\textbf{0.005} }
&\multicolumn{1}{c}{\textbf{0.01} }
&\multicolumn{1}{c}{\textbf{0.05} }
&\multicolumn{1}{c}{\textbf{0.1} }
&\multicolumn{1}{c}{\textbf{0.5} }
&\multicolumn{1}{c}{\textbf{1} } \\
\hline%第二道横线
MapExpert  &39.7&	40.0&	37.4&	33.1&	25.6	&14.8&	6.7&	1.2\\

% \midrule%第二道横线

\hline%第四道横线
\end{tabular}
\caption{Robustness experiments on expert router distribution, illustrating the significant impact of disturbances on model performance. All results are conducted with 20 training epochs.}
% }
% }
\label{table8}
\end{table}

\section{Appendix C}
\subsection{Qualitative Results} 
We present two additional qualitative experiments to further substantiate the superior performance of our model. The detailed results are represented below.

\subsubsection{Single frame results.}
\begin{figure*}[t]
\centering
\includegraphics[width=0.95\textwidth]{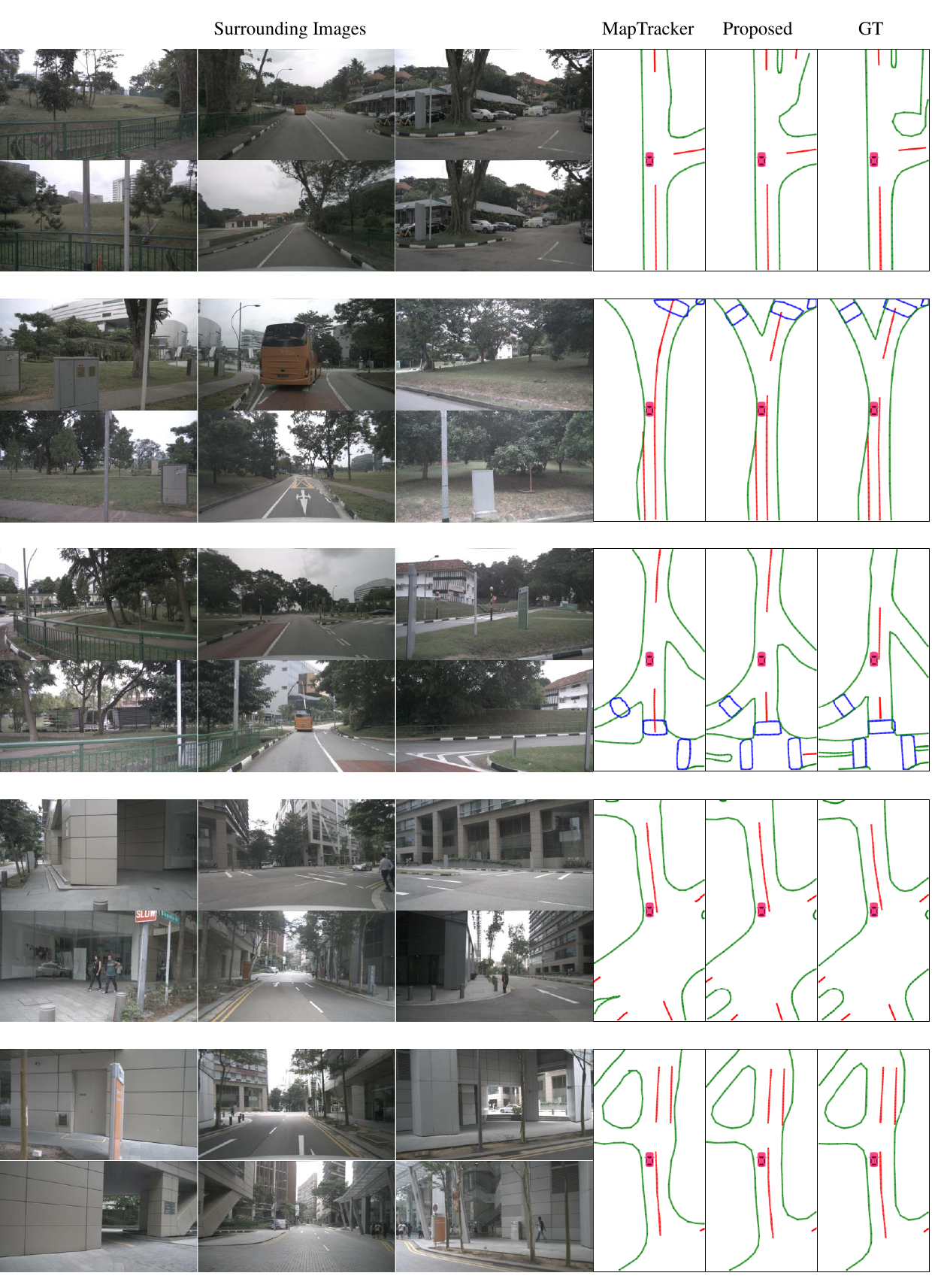} % Reduce the figure size so that it is slightly narrower than the column.
\caption{Additional qualitative results on the nuScenes dataset.}
\label{fig5}
\end{figure*}

\begin{figure*}[t]
\centering
\includegraphics[width=0.95\textwidth]{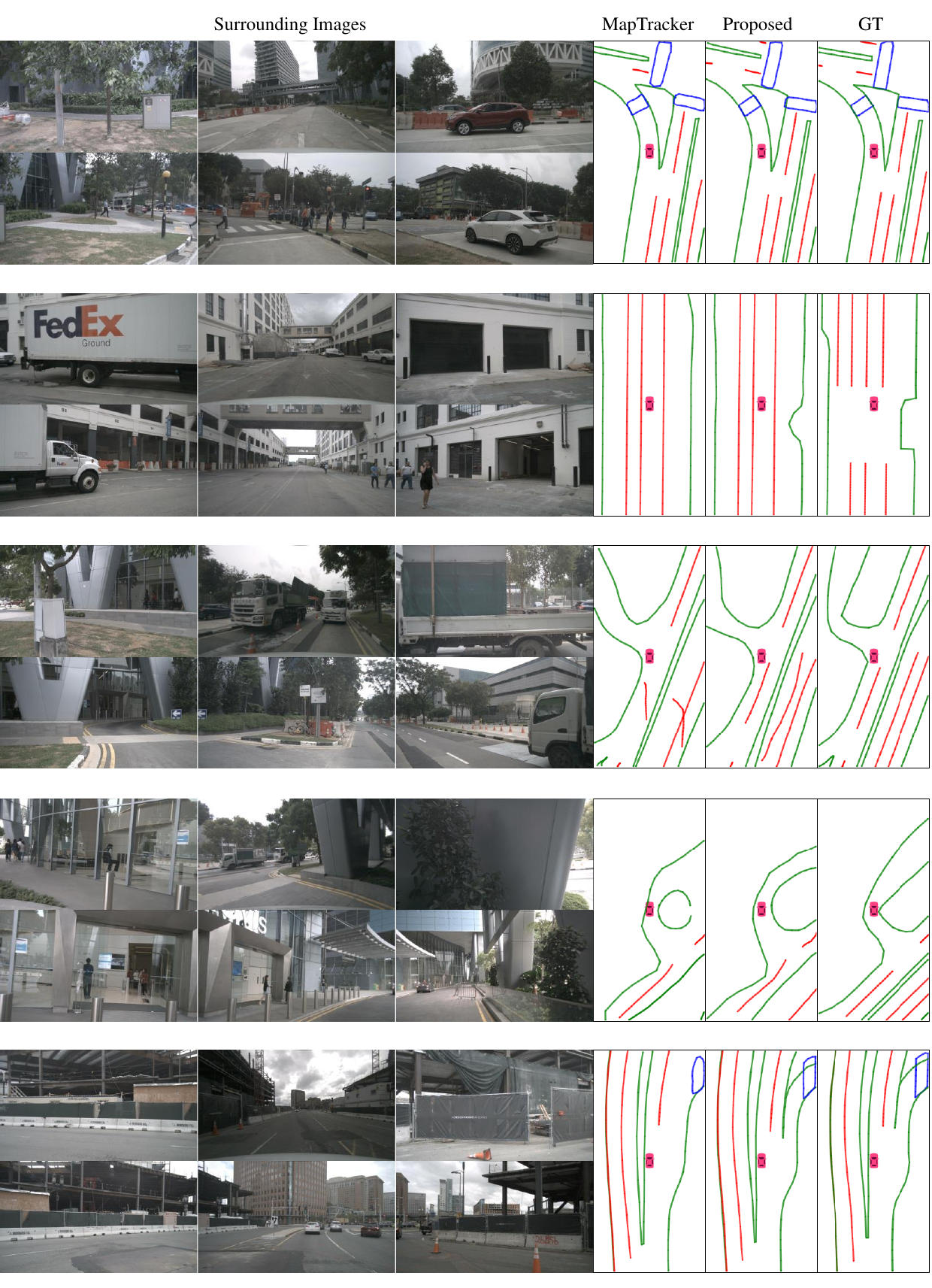}
\caption{Additional qualitative results on the nuScenes dataset.}
\label{fig6}
\end{figure*}

Additional single frame qualitative results on the nuScenes dataset are presented in Figure \ref{fig5} and Figure \ref{fig6}. Here, each group is composed of six surrounding images, MapTracker’s predictions, and MapExpert’s predictions. As you can see, our method predicts overall more accurate map elements with fewer false negatives and false positives.

\subsubsection{Scene results.}
\begin{figure*}[t]
\centering
\includegraphics[width=0.95\textwidth]{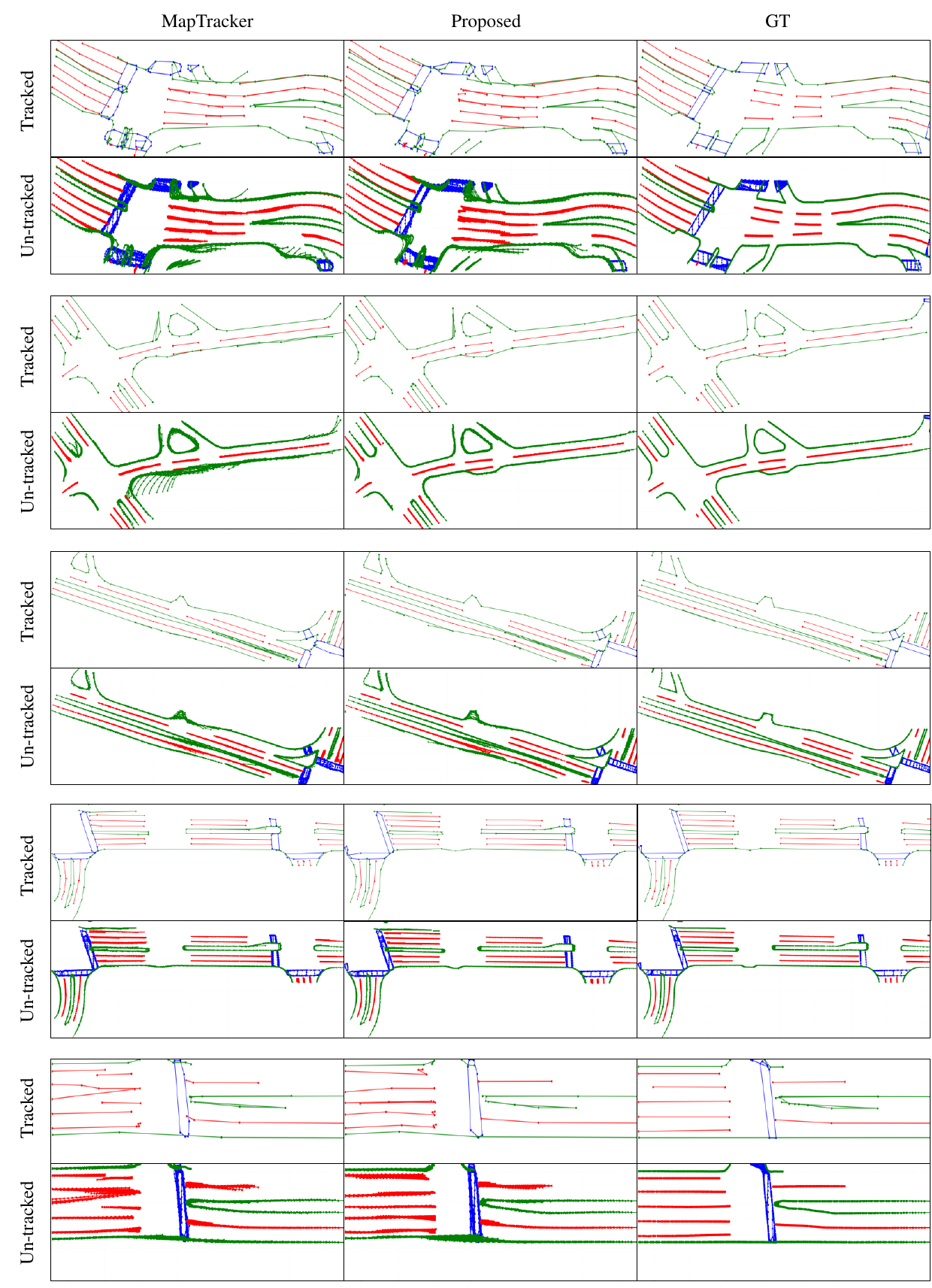} % Reduce the figure size so that it is slightly narrower than the column.
\caption{ Scene map construction over multiple frames.}
\label{fig7}
\end{figure*}

We conduct another qualitative experiment. by leveraging tracking IDs to fuse the historical construction results of each scene. As shown in Figure \ref{fig7}. The first row of each group is the merged map, which is linked by our tracking ID. The second row is the accumulated map. Comparing each row of each example, our method provides cleaner and more consistent results, particularly in challenging scenes such as crossroads and intersections. 

\section{Other Supplementary Material}
\subsection{Loss}
The total training loss is composed of four components, it is defined as:
$$ L_{total} = \beta_{1} L_{cls} + \beta_{2} L_{reg} +\beta_{3} L_{seg} +\beta_{4} L_{expert-balance} \eqno{(11)} $$
where $\beta$ is hyper-parameters, $L_{cls}$ is classification matching cost, we utilize Focal Loss. The polyline-wise matching cost $L_{reg}$ is calculated using SmoothL1 Loss. To facilitate convergence, we also incorporate a segmentation loss $L_{seg}$, which is another Focal Loss. The last is the expert balance loss, as we mentioned in Eq. 8.

\subsection{Metrics}
Our experiments adopt average precision (AP) to assess the map construction quality, and Chamfer distance is used for matching predictions with ground truth labels. We follow existing works to calculate $AP_{d}$  values with $d \in \{0.2, 0.5, 1.0\}m$ and $d \in \{0.5, 1.0, 1.5\}m$ for pedestrian crossings, lane dividers, and lane boundaries; and report the mean average precision (mAP).

\subsection{Hyper-parameters}

For the training process, we utilize AdamW optimizer with a gradient clipping norm of 2.0 (maximum norm is 35). The learning rate is managed using a CosineAnnealingLR schedule, which includes a linear warm-up period during the first 500 steps. Additionally, we apply data augmentation to the surrounding images, which enhances model robustness.

\subsection{Future Work}

While our proposed MapExpert framework demonstrates significant advancements in HD map construction and processing, several avenues remain for further exploration and improvement. Future work will focus on the following key areas:

\subsubsection{Optimization for Real-Time Applications:} Currently, MapExpert leverages powerful GPU hardware, which may not be practical for real-time applications. Future work will focus on optimizing the computational efficiency of the model and exploring lightweight architectures that enable real-time HD map generation on devices with limited computational resources.
\subsubsection{Integration of Standard-Definition (SD) Maps:} Incorporating SD maps to improve overall performance is a promising direction. Future research could explore methods for integrating SD map updates, allowing for continuous refinement of map elements based on SD maps and adapting to changing conditions.
\subsubsection{Simplified Feature Extraction:} The current encoder in MapExpert is resource-intensive due to its focus on BEV feature extraction. Future work could explore the possibility of directly extracting map features from image features using sparse feature extractors, potentially reducing the computational load and improving efficiency.

By addressing these areas, future research can further advance the capabilities of MapExpert, making it a more robust, versatile, and efficient tool for HD map construction and autonomous navigation.

\end{document}